\documentclass[runningheads]{llncs}
\usepackage[T1]{fontenc}
\usepackage{graphicx,verbatim}

\usepackage{amsfonts} 
\usepackage{amsmath}
\usepackage{multirow}
\usepackage{tabularx}
\usepackage{algorithm}
\usepackage{algpseudocode}
\usepackage{svg}
\usepackage{hyperref}
\usepackage{xcolor}
\usepackage{adjustbox}
\usepackage[normalem]{ulem}
\useunder{\uline}{\ul}{}

\begin{document}
\title{Pseudo-D: Informing Multi-View Uncertainty Estimation with Calibrated Neural Training Dynamics}

\author{Ang Nan Gu\orcidID{0000-0001-8926-2397} \inst{1} \and
Michael Tsang\inst{2} \and
Hooman Vaseli\inst{1} \and
Purang Abolmaesumi\inst{1} \and 
Teresa Tsang\inst{2}
}
\authorrunning{A. Gu et al.}
\institute{Department of Electrical and Computer Engineering, University of British Columbia, Vancouver, Canada \\
\email{guangnan@ece.ubc.ca}\\
 \and
Division of Cardiology, Vancouver General Hospital, Vancouver, Canada
\footnote{T. Tsang and P. Abolmaesumi are joint senior authors.}
}

\begin{comment}
\author{Anonymized Authors}  %% Added for anonymized MICCAI 2025 submission
\authorrunning{Anonymized Author et al.}
\institute{Anonymized Affiliations \\
    \email{email@anonymized.com}}    
\end{comment}

\maketitle              % typeset the header of the contribution

\begin{abstract}
%Computer-aided diagnosis systems face two key challenges: making predictions from medical images that may contain noise or imaging artifacts, and integrating multiple competing sources of information into a single robust decision. Existing neural network-based methods can struggle under high uncertainty and incomplete information, leading to excessive confidence and overfitting. Current medical datasets are also limited by one-hot labeling, which fail to capture any uncertainty observed during the diagnosis process and may not include inter-rater variability. We propose to bring uncertainty back to the label space by leveraging neural network training dynamics (NNTD). By analyzing training history through aggregation and calibration, we extract meaningful insights about the difficulty of each training sample. We introduce a novel label augmentation approach to generate uncertainty-aware pseudo-labels, which can be applied to any network architecture to improve uncertainty estimation. We validate our method on a challenging echocardiography classification dataset and show that it outperforms existing specialized methods in calibration, selective classification, and multi-view fusion tasks. Our source code is publicly available at: [link].
Computer-aided diagnosis systems must make critical decisions from medical images that are often noisy, ambiguous, or conflicting, yet today’s models are trained on overly simplistic labels that ignore diagnostic uncertainty. One-hot labels erase inter-rater variability and force models to make overconfident predictions, especially when faced with incomplete or artifact-laden inputs. We address this gap by introducing a novel framework that brings uncertainty back into the label space. Our method leverages neural network training dynamics (NNTD) to assess the inherent difficulty of each training sample. By aggregating and calibrating model predictions during training, we generate uncertainty-aware pseudo-labels that reflect the ambiguity encountered during learning. This label augmentation approach is architecture-agnostic and can be applied to any supervised learning pipeline to enhance uncertainty estimation and robustness. We validate our approach on a challenging echocardiography classification benchmark, demonstrating superior performance over specialized baselines in calibration, selective classification, and multi-view fusion.% Our method improves model reliability without requiring changes to network architecture or training routines. %Our code will be available at: \url{https://github.com/an-michaelg/RT4U}.
% Authors must provide keywords and are not allowed to remove this Keyword section.

\end{abstract}
\section{Introduction}

Medical image-based diagnosis faces two key challenges. First, it is safety-critical, where diagnostic errors can have serious consequences. Second, image acquisition is inherently imperfect. For example, ultrasound imaging depends heavily on the sonographer’s skill and patient-specific factors, making it difficult to capture images at the optimal angle consistently. These limitations can lower image quality and, in turn, compromise diagnostic accuracy. 
% How can we mitigate this?
As a result, evaluating these systems requires more than measuring classification accuracy; it must also account for the model’s uncertainty estimates. In challenging cases or when image quality is poor, the system should know to abstain from making a prediction and instead refer the case to a human expert. This capability is assessed through the task of selective classification.
Moreover, arriving at a diagnosis often requires integrating information from multiple sources, such as different imaging modalities or several views of the same anatomy. Crucially, the fusion process must account for the varying degrees of uncertainty associated with each source. This capability is evaluated through the task of multi-view fusion.

% contributions
We propose a method to enhance selective classification and multi-view fusion by improving uncertainty estimation (UE). We focus on aleatoric uncertainty (AU), the irreducible uncertainty caused by incomplete or ambiguous input data. Although AU is less common in traditional vision tasks, it is critical in medical imaging, where anatomical features can be obscured or ambiguous because of patient variability, the imaging process, and modality-specific factors. %The method is agnostic to both architecture and loss function, making it readily adaptable to existing approaches.

A core challenge in estimating AU is the lack of ground-truth labels that faithfully capture uncertainty. Classification datasets typically provide a single, definitive label, even though expert assessments often reflect borderline cases or ambiguity. Moreover, the common use of ``one-hot'' labels fails to convey the nuanced reasoning needed in uncertain cases. To achieve this, the model’s confidence should be better aligned with task difficulty. %To address this, we propose augmenting the label distribution dynamically based on the difficulty of the case. This allows the model to learn a wider range of uncertainty.
% something about the usage of NNTD
%We leverage neural network training dynamics (NNTD) to generate pseudo-labels with varying levels of uncertainty. NNTD can be interpreted as an ensemble constructed using different phases of the training process. By analyzing the mean and variance of these predictions, studies have shown successful use of NNTD for detecting mislabeling, classification and uncertainty estimation.

We leverage Neural Network Training Dynamics (NNTD) to generate pseudo-labels that quantify uncertainty based on how confidently and consistently the model learns each sample during training. Rather than relying on a fixed label, we track the model’s evolving predictions across epochs and treat this trajectory as a measure of sample difficulty. NNTD-based methods have proven effective in detecting label noise \cite{pleiss2020aum,seedat2022dataiq}, improving classification \cite{rabanser2022selective}, and producing more reliable uncertainty estimates \cite{gu2024rt4u}.%, because the approach implicitly ensembles predictions from different training stages.

%We improve upon previous approaches by novel augmentation function which accounts for both training and validation performance.
%This leads to more reliable AU estimation and improves the model's trustworthiness under ambiguous scenarios.
%Our method, \textit{Pseudo-D}, combines NNTD information from both the training and validation sets to better calibrate pseudo-labels at the sub-class level. This is helpful when certain sub-classes are harder to distinguish than others. We evaluate \textit{Pseudo-D} on a multi-view ultrasound dataset for aortic stenosis (AS) classification, a challenging task that requires integrating aortic valve information from multiple scanning planes. Patient-specific variation in acquisition further complicates the task by occasionally yielding poor valve visualization.% (see Figure~\ref{fig:visual_abs}).
We propose \textit{Pseudo-D}, a novel technique which combines NNTD information from both the training and validation sets to created pseudo-labels which are calibrated at the sub-class level. This is particularly useful when certain sub-classes are harder to distinguish than others.
We evaluate Pseudo-D on a challenging multi-view ultrasound dataset for aortic stenosis (AS) classification, which requires integrating information from multiple scanning planes and handling patient-specific variability in image quality.
%We evaluate \textit{Pseudo-D} on a multi-view ultrasound dataset for aortic stenosis (AS) classification. The task is challenging, requiring integration of aortic valve information from multiple scanning planes. Furthermore, patient-specific variation in acquisition can produce suboptimal valve visualization. % (see Figure~\ref{fig:visual_abs}).
We demonstrate that training with \textit{Pseudo-D} improves uncertainty estimation in standard deep learning classifiers, and outperforms specialized methods on selective classification and multi-view fusion tasks. Compared to existing approaches, our method better aligns model uncertainty with input-specific factors like image quality and anatomical visibility. Furthermore, \textit{Pseudo-D} is agnostic to model architecture and requires minimal changes to integrate to existing customized workflows.

% We show that: 
% - Training with Pseudo-D improves the uncertainty estimation of generic deep learning-based classifiers.
% - We also demonstrate superior performance to specialized approaches for selective classification and multi-view fusion.
% - Our qualitative assessment shows that Pseudo-D allows the uncertainty of the model to more closely align with input-specific factors such as image quality and anatomical visibility.

% In addition:
% - Compared to other approaches, implementing pseudo-D requires minimal changes to any existing machine learning workflow. It is agnostic to architecture, making it easily integrable with other custom workflows.

%Our contributions are as follows:
%\begin{itemize}
%    \item We propose , a novel data augmentation method which uses the network's training history to add a variable degree of uncertainty based on the difficulty of individual samples.
%    \item We demonstrate the efficacy of RT4U+ as a model-agnostic method for improving selective prediction and fusion on a challenging multi-view echocardiography AS classification task.
%    %\item We present a review of NNTD-based methodology for uncertainty estimation and data quality assessment.
%\end{itemize}

\begin{figure}[t]
    \centering
    \includegraphics[width=\linewidth]{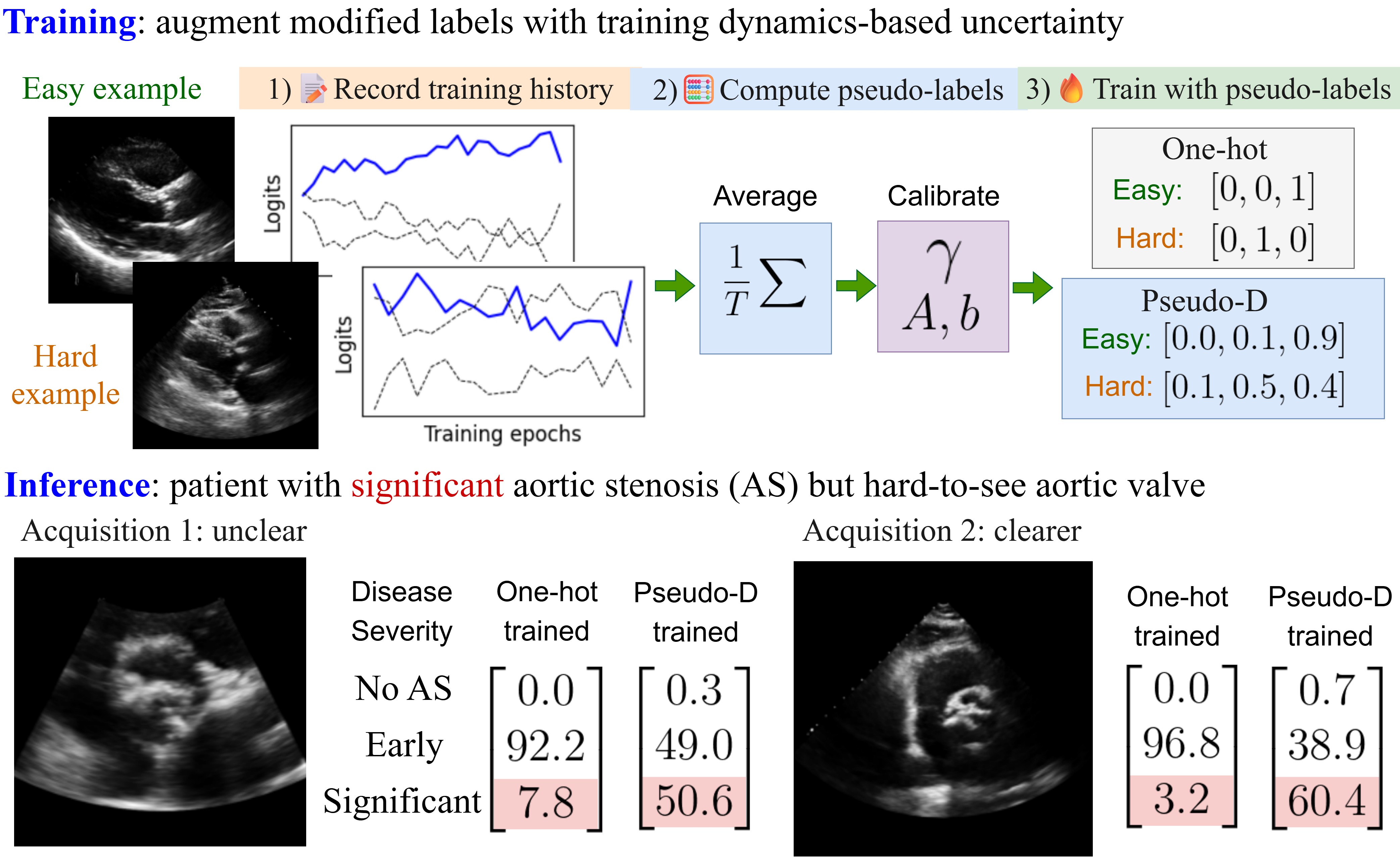}
    \caption{We augment the training phase by first recording the history of predicted logits. The magnitude of the correct class logit (shown in blue) relative to other classes (in black) varies with task difficulty. We use the training history to generate pseudo-labels that align with the difficulty of each example. Our proposed training technique, \textit{Pseudo-D}, yields a model with predicted probabilities that strongly correspond to image quality. Additionally, the pseudo-labels help mitigate overfitting by assigning lower confidence values to difficult training examples.}
    \label{fig:visual_abs}
\end{figure}

\section{Related Works}

\subsubsection{Selective Classification.}

The task of selective classification (SC), or prediction with a ``reject'' option, was initially extended to deep learning by Geifman et al.~\cite{geifman2017selective}. DeVries and Taylor~\cite{devries2018abstention} proposed explicitly learning uncertainty as an additional output of the model, using a modified loss function. 
Rabanser et al.~\cite{rabanser2022selective} suggested using model checkpoints from different training epochs to form an ensemble for SC. Huang et al.~\cite{huang2020selfadaptive} explored how label augmentation can improve SC, and Feng et al.~\cite{feng2023towards} showed that the softmax response can out-perform specialized scoring functions in existing SC approaches.
%In the context of medical imaging, Bernhardt et al.~\cite{bernhardt2022failuredetection} benchmarked various techniques for detecting misclassifications. Surprisingly, they found that the softmax score, despite its simplicity, performed best across multiple datasets.

\subsubsection{Multi-view Fusion.}

\begin{comment}
Combination of percentages in late fusion, combination of logits, DST, using attention mechanisms like multiple instance learning. The explicit linkage of uncertainty estimation with performance in late fusion techniques.
\end{comment}

%The Dempster-Shafer Theory (DST) is a framework for modeling uncertainty arising from conflicting or incomplete evidence~\cite{shafer1976dst}. Deep learning models have also utilized Demptster-Shafer theory to quantify aleatoric and epistemic uncertainty~\cite{sensoy2018evidential,malinin2018predictive}.
Multi-view fusion involves the combination of classifier predictions, where each prediction stems from a unique view of the same underlying object. The fusion of predicted probabilities can be through averaging~\cite{satopaa2014combining}, majority voting~\cite{morvant2014majority}, or by learned weighting scheme~\cite{wang2019multiviewhuman}. Zhang et al.~\cite{zhang2023qmf} establishes a theoretical link between uncertainty estimation and multi-view classification performance for logit-based approaches.
%Dempster’s Rule of Combination (DRC) provides a principled approach to merging predictions from multiple sources into a more reliable estimate~\cite{osti2002combination}. The individual predictions are interpreted as belief masses~\cite{shafer1976dst} instead of logits and can be learned differentiably via Evidential Neural Network (ENN)~\cite{sensoy2018evidential}.
Evidential Neural Networks (ENNs)~\cite{sensoy2018evidential} are trained to output belief masses instead of logits. Belief masses can be aggregated using Dempster's Rule of Combination, a mathematically rigorous method for combining multiple predictions~\cite{han2022tmc}.

%\subsection{Applications of Neural Network Training Dynamics}

%\begin{comment}
%Applications of NNTD in the label space for data screening, UE, selective prediction, data augmentation…
%\end{comment}

\subsubsection{Aortic Stenosis Severity Classification.}
\begin{comment}
Aortic Stenosis is a heart valve disease where the blood flow through the aortic valve is restricted. The clinical standard for Aortic Stenosis diagnosis is based on the blood flow volume through the left ventricle outflow tract. The volume is typically derived from spectral Doppler. Due to the sensitivity of the diagnosis to Doppler measurement, as well as the lack of spectral Doppler on more recent, lightweight ultrasound devices, F proposed prediction of AS severity through B-mode ultrasound inspection by human clinicians.

A body of work featured AI-based AS classification using B-mode. Of those, A B C proposes using one view, D E F proposes using multiple views obtained from retrospective exams in a passive environment. The evidence from several single-view models is combined using majority vote or averaging. G uses multiple instance learning to read images from an entire examination simultaneously, yielding superior accuracy, but with the downside increased computation and requiring multiple images for prediction. Our work is the first to consider AS classification in an active environment, which resembles a real-world screening procedure more.
\end{comment}

Aortic stenosis (AS) is a heart valve disease characterized by restricted blood flow through the aortic valve. The clinical standard for AS diagnosis relies on measuring blood flow volume through the left ventricular outflow tract, typically derived from spectral Doppler~\cite{aha_2020}. However, Doppler-based diagnosis is sensitive to measurement variability~\cite{thaden2015doppler_sensitivity,pibarot2012improving_sensitivity} and is often unavailable on newer, lightweight ultrasound devices~\cite{gulivc2016pocket}. Recent clinical works~\cite{abe2021screening,nemchyna2021validity} proposed assessing AS severity through B-mode ultrasound interpretation by human clinicians.
AI-based AS classification using B-mode ultrasound has gained traction: works~\cite{dai2023identifying,guo2021predicting} classify using single-image or video inputs, while~\cite{ahmadi2023transformer,holste2023severe,huang2021tmed1,krishna2023us2ai,vaseli2023protoasnet} and~\cite{wesslerAutomatedAS2023} utilize multiple views from retrospective exams, combining predictions via majority voting or averaging of predicted confidence. Huang et al.~\cite{huang2024mil} uses multiple-instance learning to learn the importance of each image.

\section{Methodology}

\subsection{Background: Construction of pseudo-labels via NNTD}
\begin{comment}
    Data in the training set vary in terms of their difficulty to be fit by the predictive model, whether that is due to the difficulty of the task (ie. distinguishing between two similar classes is geniunely difficult even in optimal imaging conditions), the presence of input-dependent noise (eg. introduction of imaging artefacts, lack of visibility), or labeling error. Empirical observations (cite) show that the rate at which a training sample is learned depends on this difficulty. Thus, analysis of the neural network training dynamics provide insight into the difficulty of training examples. One of the ways to use the training history is to conduct training once to collect information on each example. Based on how quickly each example fit their label, a new pseudo-label can be created which more adequately reflect the difficulty of the task. (equation here). These new labels, which include a finite degree of uncertainty based on their difficulty, can be used to train another model. This strategy is also effective in data analysis for quickly identifying potentially mislabeled (SEAL) or ambiguous (RT4U) examples within the training set.
\end{comment}

Training samples $\mathcal{X}_{\text{train}} = {(x_0, y_0), \ldots, (x_N, y_N)}$ vary in how difficult they are for a model to learn. This difficulty may stem from intrinsic task complexity (e.g., distinguishing visually similar classes), data-related issues (such as noise, occlusions, or imaging artifacts), or label noise. %The rate at which a neural network learns to predict the correct label $y_i$ for a sample $x_i$ often reflects this underlying difficulty.
The model’s evolving predictions $f(x_i)$ over $T$ training epochs provide insight into the difficulty of each data point. For easy examples, the logit corresponding to the ground truth class tends to dominate consistently across epochs. In contrast, for difficult or mislabeled examples, the logits for different classes often remain uncertain or competitive. We can compute pseudo-labels $y_i' \in [0, 1]^C$ which represent the model’s average class confidence over time:

\begin{equation}\label{eqn:pseudo_naive}
y_i' = \frac{1}{T} \sum_{t=1}^{T} \sigma(f_{t}(x_i)), \quad \quad x_i \in \mathcal{X}_{\text{train}},
\end{equation}
where $f_t(x_i)$ denotes the model logits at epoch $t$, and $\sigma(\cdot)$ is the softmax function applied over class outputs.

These pseudo-labels encode the uncertainty related to sample difficulty, enabling more robust learning. Prior work has shown that training on such soft targets improves resistance to label noise~\cite{chen2021seal,huang2020selfadaptive} and enhances uncertainty estimation~\cite{gu2024rt4u}. Training on $y'$ can be viewed as a form of knowledge distillation from a training-dynamics-based ensemble, capturing model behavior across epochs. Crucially, these pseudo-labels are architecture-agnostic and can be used to train any downstream model on the same task.

\subsection{Calibrated pseudo-labels}
\begin{comment}
    A weakness with the pseudo-labels in the previous method is that the confidences may not necessarily be calibrated. In essence, if a neural network is optimized to fit the pseudo-labels, the probabilties it produces may still be over- or under-confident. To counteract this, we propose a new method of creating the pseudo-labels which take into account both the training and validation history. (eqn where this is that).
    Where gamma is a temperature scaling coefficient that minimizes the NLL with respect to the validation set (eqn). We denote this method as pseudo (temp). Temperature scaling has been applied to many neural network prediction as a post-processing step to mitigate over-confidence. However, the use of a single coefficient for logits of all classes does not account for variations in the accuracy of predictions for each subclass. We leverage a more complex method, namely Dirichlet Calibration, to produce pseudo-labels which are calibrated with respect to each sub-class. (eqn where this is that).
\end{comment}

A limitation of Eqn.~\ref{eqn:pseudo_naive} is that the confidence of $y_i'$ may not be well-calibrated, i.e. even if $f(.)$ fits these pseudo-labels, the predicted probabilities might still be over- or under-confident.
To address this, we generate pseudo-labels using information from both $\mathcal{X}_{train}$ and $\mathcal{X}_{val}$. Specifically, we apply temperature scaling to the logits so that output confidences better reflect the true accuracy (Eqn.~\ref{eqn:avg_logit} - \ref{eqn:temp_scaling}):

\begin{align}
    v_i &= \frac{1}{T} \sum_{t=1}^T f_{t}(x_i), \label{eqn:avg_logit} \\
    y_i' &= \sigma(\gamma^* \ v_i), & x_i \in \mathcal{X}_{\text{train}}, \label{eqn:pseudo_logit} \\
    \gamma^* &= \operatorname*{argmin}_\gamma \ \text{CrsEnt}(\gamma \ v_j, y_j), & x_j \in \mathcal{X}_{\text{val}} ,\label{eqn:temp_scaling}
\end{align}
where $v_i \in \mathbb{R}^C$ denotes the logit vector averaged over epochs for both $\mathcal{X}_{train}$ and $\mathcal{X}_{val}$. The temperature parameter, denoted as $\gamma^*$, is chosen to minimize the negative log-likelihood (NLL), which is equivalent to minimizing cross entropy, on $\mathcal{X}_{val}$. We refer to this approach as \textit{Pseudo-T}. 

While temperature scaling is a common technique for improving model calibration, it typically applies a single global scaling factor across all classes. This uniform treatment fails to account for class-specific variability, since some classes can be inherently harder to differentiate compared to others.

To address this limitation, we adopt Dirichlet Calibration~\cite{kull2019dirichlet}, which applies class-wise scaling to the model logits. This approach learns a transformation matrix and bias that adjust each class individually, yielding pseudo-labels with improved sub-class calibration:

\begin{align}
y_i' &= \sigma(A^* v_i + b^* ), \quad \quad x_i \in \mathcal{X}_{\text{train}}, \label{eqn:pseudo_matrix} \\
A^* , b^* &= \arg\min_{A, b} \ \text{CrsEnt}(Av_j + b, y_j) + \frac{\lambda_1}{C} |b|_2^2 + \frac{\lambda_2}{C^2} |\bar{D}(A)|_2^2, \quad x_j \in \mathcal{X}_{\text{val}}. \label{eqn:matrix_scaling}
\end{align}

Here, $v_i$ denotes the model logits, $A \in \mathbb{R}^{C \times C}$ is the learned scaling matrix, and $b \in \mathbb{R}^C$ is a bias vector. Compared to simple temperature scaling (\textit{Pseudo-T}), this method introduces significantly more parameters, increasing the risk of overfitting to the validation set $\mathcal{X}_{\text{val}}$. To mitigate this, we apply regularization terms: one to penalize large bias magnitudes, and another to suppress off-diagonal entries in $A$ (denoted $\bar{D}(A)$). 
In practice, this results in a flexible yet stable calibration scheme: the diagonal entries of $A$ remain expressive, while off-diagonal interactions are dampened. We refer to this pseudo-label method as \textit{Pseudo-D}.

\section{Experiments}

\subsection{Dataset}
We use an anonymized private dataset obtained from a tertiary hospital with ethics approval. The dataset contains 2572 retrospective echo studies, acquired with Philips iE33, Vivid i, and Vivid E9 transducers.
The studies were first labeled as normal/mild/moderate/severe based on spectral Doppler measurements using AS diagnostic guidelines from \cite{bonow2006acc}. Consistent with prior methods~\cite{huang2021tmed1,wesslerAutomatedAS2023,vaseli2023protoasnet}, the moderate and severe classes are combined into a single ``significant class''. An experienced cardiologist selected parasternal long-axis and short-axis views from each study, resulting in a total of 9117 videos. We created training, validation and test using a randomized 80/10/10 split, ensuring no patients overlap across subsets. Each video was preprocessed by extracting approximately one heart cycle, removing background UI elements, and resizing to $16\times224\times224$.

\subsection{Evaluation Procedure and Metrics}
We compare models trained on the following pseudo-labels: \textit{RT4U}~\cite{gu2024rt4u} (Eqn. \ref{eqn:pseudo_naive}), \textit{Pseudo-T}, and \textit{Pseudo-D} on selective classification and multi-view fusion.
We compare the pseudo-label approaches with \textit{Vanilla}, a baseline cross-entropy approach; \textit{Abstention}~\cite{devries2018abstention}, which trains an extra network branch specializing in rejecting uncertain predictions; and \textit{TMC}~\cite{han2022tmc}, which specializes in combining probabilities from multiple sources.

In selective classification, models are provided with an option to ``reject'' the prediction $f(x)$ based on a selection function $g(x)$ and threshold $\tau$. The effectiveness of selective classification depends on both the accuracy of $f(x)$ and the sensitivity of $g(x)$ for identifying likely misclassifications. The coverage (Eqn.~\ref{eqn:aurc_cov}) and accuracy (Eqn.~\ref{eqn:aurc_acc}) are evaluated at different thresholds. Performance over multiple thresholds can be summarized by the Area Under Risk-Coverage Curve (AURC)~\cite{geifman2017selective} (Eqn.~\ref{eqn:aurc}):

\begin{align}
    \text{Cov}(f, g, \tau) &= |{x: g(x) > \tau}|, \label{eqn:aurc_cov} \\
    \text{Acc}(f, g, \tau) &= \frac{|{(x, y): f(x) = y, g(x) > \tau}|}{|x: g(x) > \tau|}, \label{eqn:aurc_acc} \\
    \text{AURC}(f, g) &= \frac{1}{|\mathcal{T}|}\sum_\tau \text{Acc}(f, g, \tau) *  \text{Cov}(f, g, \tau). \label{eqn:aurc}
\end{align}
We choose $g(x)$ to be tied to the softmax confidence of the predicted class, since it was shown to be successful across multiple specialized methods~\cite{feng2023towards}. Traditionally the AURC measures loss; we co-opt the metric to assess balanced accuracy due to the nature of the classification task. We pre-compute the set of thresholds $\mathcal{T}$ at the percentiles of coverage ranging uniformly from 50\% to 100\%.

We use multi-view fusion to aggregate video-level predictions to study-level. %In some studies, only one video in the study display biomarkers which indicate illness, since acquiring high-quality video of the aortic valve is challenging. Therefore, we adopt a ``worst-case'' approach to aggregate the predictions from video- to study-level. 
In some studies, only one or two videos clearly show signatures of stenosis on the aortic valve. Thus, we adopt a ``worst-case'' aggregate strategy. If all videos in the study are predicted as normal, we average the softmax probability. Otherwise, we average only over the subset of videos predicted as abnormal.

In terms of metrics, we measure the accuracy (ACC), expected calibration error (ECE), and mean average error (MAE) with class 0 as normal, 1 as early, and 2 as significant. To account for class imbalance, we report balanced metrics by computing each metric per class and then averaging the results.

\subsection{Implementation Details}
For both pseudo-label generation and fitting, we model $f(.)$ using R(2+1)D~\cite{tran2018r2p1d} with Kinetics400 initialization, trained with ADAM, learning rate 1e-4 for 20 epochs. We augment the input via random rotation of $\pm 15 ^\circ$ and cropping with ratio $0.7$.
To compute $y_i'$, we save the logits from $\mathcal{X}_{train}$ and $\mathcal{X}_{val}$ every epoch; we tune temperature parameters $\gamma$, $A$ and $b$ using SGD with learning rate $0.01$ and $\lambda_1 = \lambda_2 = 1$. %The source code for experiments will be made available.

\begin{table}[t]
\centering
\caption{Evaluation over the test set of the aortic stenosis classification dataset. Each study consists of multiple ultrasound videos of the same patient. Models predict at the video-level, and predictions are fused into the study-level. We report metrics computed separately for each class, then averaged. Best and second-best results are \textbf{bolded} and {\ul underlined}, respectively.}
\begin{tabular}{l|cccc|cccc}
\hline
           & \multicolumn{4}{c|}{Video-level}         & \multicolumn{4}{c}{Study-level}           \\ 
Method     & MAE $\downarrow \ $   & ACC $\uparrow \ $  & ECE $\downarrow \ $  & AURC $\uparrow$   & MAE $\downarrow \ $   & ACC $\uparrow \ $    & ECE $\downarrow$   & AURC $\uparrow$    \\ \hline \hline
Vanilla    & .257       & .765       & .170      & .804          & .224          & .790          & .138      & .852       \\
Abstention~\cite{devries2018abstention} & .250       & .769       & .130      & {\ul .820}    & .216          & .794          & .094       & {\ul .881} \\
TMC~\cite{han2022tmc}   & .252       & .771       & .213      & .746          & .205          & .808          & .163      & .777       \\
RT4U~\cite{gu2024rt4u}   & .246       & .773       & .105      & .804          & \textbf{.171} & \textbf{.836} & {\ul .073} & .877       \\
Pseudo-T (ours)  & {\ul .234} & {\ul .786} & {\ul .097} & \textbf{.838} & .192          & .808          & .089       & .874       \\
Pseudo-D (ours) & \textbf{.220} & \textbf{.787} & \textbf{.096} & .793 & {\ul .180} & {\ul .820} & \textbf{.071} & \textbf{.885} \\
\hline
\end{tabular}
\label{tab:quant_results}
\end{table}

\subsection{Results and Discussion}
% Quantitative performance of pseudolabel methods
% Coefficients for Dirichlet scaling matches with the difficulty of each class
\begin{comment}
- Pseudo-D method performs well on most metrics in terms of video- and study-level.
- Pseudo-T method is beneficial, but due to the lack of class-specific calibration, it is not as good as Pseudo-D.
- Abstention method which is specialized for selective prediction is good for AURC.
- TMC method, while close to others in accuracy, is not as calibrated as others, since it was designed only for decision aggregation.
- Overall, pseudo-label based methods can reduce overfitting to potentially noisy or uninformative samples in the training set
- On top of that, Pseudo-D accounts for the fact that confusion between classes are typically heterogeneous in medical datasets. In this case, it is easier to differentiate between normal/early compared to early/significant.
- Temperature scaling applies the same coefficient to all logits, in contrast, Dirichlet scaling can affect each class differently, and therefore change the ordering of classes
- Here we show the gamma from our experiments and A, b. The off-diagnonal and bias parameters are close to zero. The largest practical difference is the coefficients for each class.
\end{comment}

Table \ref{tab:quant_results} compares the pseudo-labeling and specialized methods at both the video- and study-level. Overall, \textit{Pseudo-D} performs best across most evaluation metrics. %Its class-specific calibration is important in medical data where some classes are easier to tell apart than others—for example, it's easier to distinguish normal from early disease than early from significant. 
\textit{Abstention} is effective for selective classification based on AURC, but under-performs on other metrics. \textit{TMC} has similar accuracy and MAE to other methods, but its predictions are less well-calibrated, as the method was designed primarily for aggregation rather than uncertainty estimation.

The pseudo-label based methods may be performing better on noisy imaging modalities such as ultrasound because they reduce overfitting to difficult training examples. The difference between these methods is how the output probabilities are scaled. \textit{Pseudo-T} improves over the baseline, but its lack of class-wise calibration is limiting. This is due to the heterogeneity in class confusion. In this instance, it is easier to distinguish between normal and early than between early and significant aortic stenosis disease. The learned calibration parameters $\gamma^*$, $A^*$ and $b^*$ show that off-diagonal terms and biases are near zero, with the most notable differences coming from the class-specific scaling factors, %(Eqn. \ref{eqn:learned_calibration}).

\begin{align*}
    \gamma^* = [0.698], \quad 
    A^* = \begin{bmatrix}
        0.944 & 0.070 & -0.064\\
        -0.083 & 0.621 & 0.085\\
        0.061 & -0.056 & 0.591
    \end{bmatrix}, \quad 
    b^* = \begin{bmatrix}
        -0.026 \\ 0.003 \\ 0.029
    \end{bmatrix}. %\label{eqn:learned_calibration}
\end{align*}

% Qualitative performance
\begin{comment}
    We showcase an example of inference on a difficult study in figure 1. Here, the patient is diagnosed with significant AS. However, the ultrasound acquisition procedure for this patient was difficult and the videos are low quality. In the first video, the lesion is unclear. The vanilla approach, which is trained on one-hot label, is overconfident in predicting the severity. In contrast, Pseudo-D are informed by the difficulty (based on training history) and calibration against the validation set. While it is able to rule out the possibility of no disease, it remains uncertain on the severity. When given the clearer image, the confidence for the significant class increases.
\end{comment}

Figure~\ref{fig:visual_abs} presents an inference example from a particularly challenging case with significant AS. The ultrasound acquisition was difficult, resulting in low-quality videos. In acquisition 1, the valve is not clearly visible. The baseline model, trained using one-hot labels, is overconfident despite the poor image quality. In contrast, the model trained with \textit{Pseudo-D} remains appropriately uncertain about the severity. For acquisition 2, the model’s confidence increases accordingly due to the improved visual clarity.

% Misclassification detection and attribute shift mitigation (bicuspid valve)
\begin{comment}
    We also look at the detection of aortic stenosis when the a confounding factor is added to the disease, the bicuspid valve. The aortic valve is typically tricuspid, meaning it consists of three leaflets. Bicsupid aortic valve is a congenital heart defect causing two of the leaflets to be fused. A subset of our studies (14.1 percent) are from patients with bicuspid valve. The unique geometry of bicuspid valve, plus the fact that they are the minority group both make these cases harder to correctly classify. Thus, uncertainty estimation is important, such that the misclassifications correlate with a high degree of uncertainty.
    In figure 2 we plot the distribution of confidence for each of the subgroup, as well as the average confidence for the subgroup. Ideally, we maximize the separation of the correct and incorrect subgroups. In addition, since the algorithm makes more errors on, bicuspid subgroups, on average the confidence of the bicuspid subgroup is lower. We see that both pseudolabel methods improve the separation between the subgroups, and improve the correlation of the uncertainty with the actual probability of error.
\end{comment}

\begin{figure}[t]
    \centering
    \includegraphics[width=\linewidth]{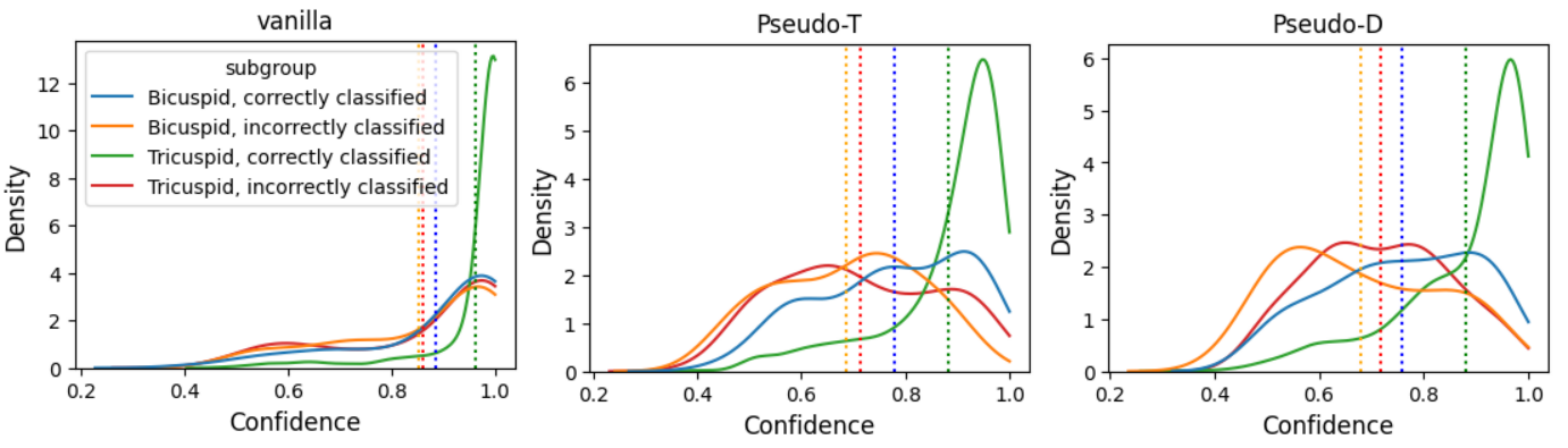}
    \caption{Density plot of predicted confidence scores, grouped by presence of bicuspid aortic valve and prediction correctness. Dotted lines show the average confidence for each subgroup. Models trained with pseudo-label methods show stronger separation, meaning cases harder to classify correctly are more identifiable through uncertainty.}
    \label{fig:bicuspid}
\end{figure}

We evaluate classification for the subset of patients with bicuspid aortic valve (BAV). The aortic valve is normally tricuspid, consisting of three leaflets. BAV is a congenital defect where two leaflets are fused, consisting of 14.1\% of studies in our dataset. These cases are harder to classify due to their under-representation in the dataset and atypical valve morphology. Accurate uncertainty estimation is key to ensuring that misclassifications are flagged with high uncertainty.
In Figure~\ref{fig:bicuspid}, we show the distribution of model confidence scores for each subgroup. Ideally, the confidence distributions for correct and incorrect predictions should be well separated. Since models perform worse on BAV cases, we also expect lower average confidence for this subgroup. Both pseudo-labeling methods improve subgroup separation and better align uncertainty with actual error likelihood.

\section{Conclusion}

\begin{comment}
- We introduced our method for approximating the difficulty of training data samples and producing pseudo-labels which reflect the unique difficulty of each case. We tie this notion of difficulty with the uncertainty estimation of the model and show that the resultant models improve in terms of selective prediction and multi-view fusion, two tasks where uncertainty estimation is important.
- In general, medical imaging datasets have more frequent issues in terms of data quality (due to disparities in image quality) and label disagreement (due to inter-clinician variability) compared to typical computer-vision datasets, the methods which are effective in computer vision may not carry over one-to-one.
- We encourage looking at more data-centric methods for improving uncertainty estimation. The variation in training dynamics can provide useful information for data quality and uncertainty, even when only one set of ground truth is available. Future work may combine training dynamics-informed methods and other approaches of estimating aleatoric uncertainty.
\end{comment}
We introduced a method for approximating the difficulty of training samples via NNTD, and generating pseudo-labels that reflect the unique challenge each case presents. %By linking this difficulty to the model uncertainty, 
We demonstrated improved performance in selective classification and multi-view fusion, two tasks where reliable UE is essential. 
%Medical imaging datasets often suffer from inconsistent image quality and label inconsistency due to inter-clinician variability. As a result, approaches from traditional computer vision may not transfer directly. We advocate for more data-centric approaches to improve UE. In particular, 
NNTD can reveal valuable insights about data quality and model uncertainty, even when only a single ground truth label is available. 

However, NNTD is still limited by and may vary based on the network architecture. Future work may explore distillation via creating/fitting pseudo-labels with larger/smaller networks respectively, combining dynamics-based strategies with other methods for capturing AU, accounting for inter-clinician variability through multiple label sets, and improving the sensitivity with respect to specific classes or sub-groups. 

\section{Acknowledgements}
This work was supported in part by the UBC Advanced Research Computing Center (ARC), the Canadian Institutes of Health Research (CIHR), and the Natural Sciences and Engineering Research Council of Canada (NSERC).

This preprint has not undergone any post-submission improvements or corrections. The Version of Record of this contribution is published in: Uncertainty for Safe Utilization of Machine Learning in Medical Imaging (UNSURE) Workshop at the International Conference on Medical Image Computing and Computer-Assisted Intervention (MICCAI), Springer (2025) under the same title.

\section{Disclosure of Interests}
The authors have no competing interests to declare that are
relevant to the content of this article.

\bibliographystyle{splncs04}
\bibliography{arxiv_ready}

\end{document}